\documentclass[11pt]{article}
\usepackage[preprint]{acl}
\usepackage{times,latexsym}
\usepackage[T1]{fontenc}
\usepackage[utf8]{inputenc}
\usepackage{microtype,inconsolata}
\usepackage{amsmath,amssymb,booktabs,array}
\usepackage{algorithm,algpseudocode}
\usepackage{tikz}
\usetikzlibrary{arrows.meta,positioning,fit,calc}
\definecolor{academicblue}{RGB}{38,77,117}
\newcommand{\E}{\mathbb E}
\newcommand{\TV}{d_{\mathrm{TV}}}
\newcommand{\push}{_{\#}}
\title{Trajectory Abstraction for the Science of Language Agent Behavior}
\author{
  Tianqiang~Yan\thanks{Corresponding author.} \\
  Faculty of Information Technology, Monash University \\
  \texttt{Tianqiang.Yan@monash.edu}
}
\begin{document}
\maketitle
\begin{abstract}
Scientific studies of language agents need behavioral variables that support hypotheses across tasks and models. We formulate this research problem as learning and testing a hierarchy of trajectory abstractions. A concrete recursive procedure first measures role- and phase-indexed events, proposes temporally constrained relations, and tests their stability across conditions. It then constructs episode-level motif variables from selected relations and repeats the analysis on those variables. Explicit measurement functions connect every abstraction level to the original trajectories. Observations and randomized protocol experiments assess the resulting hypotheses, while comparisons between intervention realizations determine whether an abstraction should be retained, refined, or restricted. We derive a finite-depth bound for accepted reductions, identify protocol effects on fixed abstractions, and characterize realization disagreement and composition of abstraction error. A finite-sample test makes projected intervention consistency operational, and constructed examples illustrate motif construction and abstraction refinement. The formulation distinguishes this experimental approach from semantic taxonomies, qualitative theory induction, and behavior-model recovery. It specifies a proposed research procedure for discovering generalizable behavioral hypotheses, with literature-relative novelty assessed separately from model-relative surprise.
\end{abstract}

\section{Introduction}
Language agents communicate, invoke tools, reflect on feedback, and coordinate actions. Scientific studies of these behaviors seek explanations that remain useful across tasks and models \citep{rahwan2019machine,chen2026behavior}. Their observations are trajectories, while their conclusions concern patterns such as checking evidence before action or completing a repair after detecting an error. The scientific problem is to construct the variables in which these patterns can be stated and tested.

A useful abstraction must preserve the distinctions relevant to the question. Consider two messages: one identifies a retrieved result and the intended action; another says ``Okay.'' Both can be counted as acknowledgments, yet the shared label can conceal differences relevant to the next action. Testing those differences determines whether to retain the count, distinguish message types, or restrict the hypothesis to one protocol. This is an abstraction-construction problem within the scientific study of behavior.

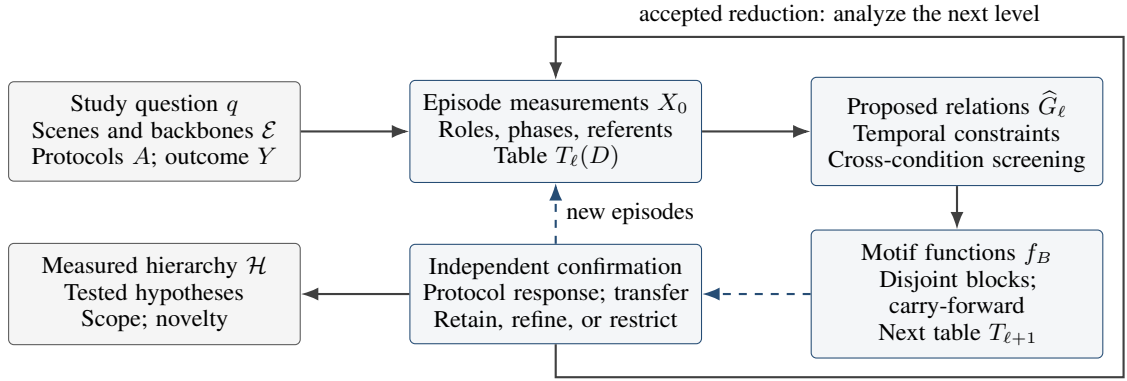
\begin{figure*}[t]
\centering
\begin{tikzpicture}[font=\small,>=Latex,
 box/.style={draw=academicblue,rounded corners=2pt,fill=academicblue!5,align=center,text width=3.5cm,minimum height=1.3cm,inner sep=5pt},
 gray/.style={box,draw=black!65,fill=black!4},
 flow/.style={->,draw=black!75,thick},test/.style={->,draw=academicblue,dashed,thick}]
\node[gray] (study) at (0,0) {Study question $q$\\Scenes and backbones $\mathcal E$\\Protocols $A$; outcome $Y$};
\node[box] (events) at (5.3,0) {Episode measurements $X_0$\\Roles, phases, referents\\Table $T_\ell(D)$};
\node[box] (graph) at (10.6,0) {Proposed relations $\widehat G_\ell$\\Temporal constraints\\Cross-condition screening};
\node[box] (motifs) at (10.6,-2.15) {Motif functions $f_B$\\Disjoint blocks; carry-forward\\Next table $T_{\ell+1}$};
\node[box] (validation) at (5.3,-2.15) {Independent confirmation\\Protocol response; transfer\\Retain, refine, or restrict};
\node[gray] (results) at (0,-2.15) {Measured hierarchy $\mathcal H$\\Tested hypotheses\\Scope; novelty};
\draw[flow] (study) -- (events);
\draw[flow] (events) -- (graph);
\draw[flow] (graph) -- (motifs);
\draw[test] (motifs) -- (validation);
\draw[flow] (validation) -- (results);
\draw[test] (validation.north) -- node[right,font=\footnotesize] {new episodes} (events.south);
\draw[flow] (validation.south) -- (5.3,-3.25) -- (12.8,-3.25) -- (12.8,1.25) -- node[above,font=\footnotesize] {accepted reduction: analyze the next level} (5.3,1.25) -- (events.north);
\end{tikzpicture}
\caption{Recursive trajectory abstraction as a scientific research procedure. Episode measurements produce a table at each level. Relations nominate temporal motifs, whose explicit measurement functions define the next table. Independent experiments test these abstractions and their hypotheses across scenes and language-model backbones.}
\label{fig:study}
\end{figure*}

Recent work already provides behavioral vocabularies, inductive analysis, and formal trace abstraction. Actonomy develops an action taxonomy; AutoTraceGT revises grounded codes and relations; ATLAS learns automata from abstract traces \citep{gao2026actonomy,lu2026autotracegt,lopezmiguel2026atlas}. ReBeCA and CAMO connect measured factors to structural analysis and experiments \citep{rebeca,camo}. These developments motivate a concrete formulation of how trajectory variables, relational hypotheses, and executable tests interact during discovery.

Our position is that higher-level behavioral explanations should be constructed through explicit trajectory functions and tested against the contrasts they claim to preserve. We operationalize this position with a recursive procedure: measure events, learn relation candidates, test cross-condition stability, construct motif variables, and evaluate their intervention responses. An accepted reduction becomes the input to the next level. The output is a hierarchy of measurements and tested hypotheses, including unresolved and rejected proposals.

The paper contributes a critical review, a computational formulation of this procedure, and analytical results supporting its design. Randomized assignments identify protocol effects on fixed abstractions even when the proposed graph is incorrect. Disagreement between realizations gives a criterion for rejecting a common-response claim. A finite-sample comparison implements that criterion, while compression and consistency bounds constrain recursion. The procedure is specified here; empirical implementation and evaluation remain future work.

\section{Related Work}
\label{sec:review}
We review primary work available through 6 October 2026, using topic searches, reference snowballing, and targeted verification of relevant papers. Appendix~\ref{app:review} summarizes the selection and comparisons.

\subsection{Behavioral Measurement and Induction}
ReAct, Reflexion, and LATS expose reasoning, actions, observations, and feedback \citep{react,reflexion,lats}. ALFWorld, WebArena, SWE-bench, and PARTNR provide repeatable tasks and verifiable outcomes \citep{alfworld,webarena,swebench,partnr}. AgentBoard analyzes intermediate progress; MAST derives failure categories; AgentDiagnose evaluates trajectory properties using rubrics and visualizations \citep{agentboard,mast,ou2025agentdiagnose}. These resources supply measurements and experimental settings.

Actonomy organizes agent actions into a reusable three-level taxonomy. AutoTraceGT performs open and axial coding, revises categories and relations, and stops according to a saturation rule \citep{gao2026actonomy,lu2026autotracegt}. AutoTraceGT therefore already models an inductive study process. Our distinction concerns recursive construction of measured motifs from relational hypotheses and confirmation through executable interventions. Category stability and saturation answer different questions from consistency of intervention responses.

Language-based measurements must connect utterances to context and action. Grounding theory, MindCraft, and embodied-dialogue analysis motivate this connection \citep{clark1991grounding,bara-etal-2021-mindcraft,worlddialogue}. Interface-changing experiments also distinguish semantic use from interface reliance \citep{gu2026interfaces}. These examples guide the definition of discriminating tests rather than treating semantic labels as explanations by themselves.

\subsection{Trajectory and Causal Abstraction}
ATLAS categorizes action--observation pairs, normalizes labels, and uses automata learning to recover Markov models; it retains concrete realizations and studies symbolic transfer \citep{lopezmiguel2026atlas}. Its recovered transition structure differs from the intervention-preserving response tested here. COAT and IRIS propose variables and evaluate relations from unstructured evidence \citep{coat,feng-etal-2025-iris}. ReBeCA combines semantic reflection patterns, structure learning, stability checks, and prompt interventions; CAMO constructs factors and performs paired simulator interventions \citep{rebeca,camo}. LLawCo tests cooperation laws as policy guidance \citep{llawco}. We build on these concrete operations while making the per-episode map at every recursive level explicit.

Causal abstraction defines consistency of interventions across descriptions \citep{rubenstein2017causal,beckers2019abstracting,beckers2020approximate}; causal representation learning concerns the variables themselves \citep{scholkopf2021toward,geiger2021causal}. Options, predictive-state representations, and bisimulation supply temporal or predictive criteria \citep{sutton1999options,littman2001predictive,givan2003equivalence}. The present construction selects its criterion according to the behavioral hypothesis and the experiments available to test it.

\subsection{Automated Scientific Research}
The AI Scientist, Co-Scientist, and AI-Researcher generate and execute research operations with scientific evaluations \citep{lu2026science,gottweis2026coscientist,airesearcher}. AutoDiscovery uses Bayesian surprise to direct hypothesis search \citep{autodiscovery}. These systems can support the proposed procedure. Its scientific outputs depend on measurements, identification, and confirmation, while surprise ranks belief changes within an explicit model \citep{itti2009surprise}. Literature-relative novelty requires an independent comparison.

\section{Problem Formulation}
\label{sec:problem}
\subsection{Study Setting}
A study specifies a question $q$, environments $\mathcal E$, joint protocol assignment $A$ with law $P_A$, an outcome function $g$, knowledge $K$, and budget $B$. In the categorical instantiation, every scored node has prespecified finite states, including $A$ and the numerical endpoint $Y$. Counts use fixed bins or truncation. Continuous endpoints require an appropriate likelihood and a fixed finite projection for categorical response tests; their original mean contrasts use integrable numerical endpoints. Environments can be indexed by scene and language-model backbone, $e=(s,m)$. Task, agent composition, interfaces, and persistent memory are fixed or included among the study factors.

An episode has trajectory $\tau$ containing timestamped observations, declared plans, messages, tool calls, actions, and execution feedback. Its outcome is $Y=g(\tau)$. Data are
\begin{equation}
D=\{(e_u,A_u,\tau_u,Y_u)\}_{u=1}^{N}.
\end{equation}
Assign complete protocol policies randomly before each experimental unit starts. A unit is an episode or a cluster with a prespecified endpoint and weighting; there is no unmodeled interference or carry-over between units. Background strata are preassignment. Confirmation units are conditionally iid within each tested environment and arm, given completed construction. Successful and failed units are retained.

\subsection{Discovery Target}
The target is a hierarchy
\begin{equation}
\mathcal H=\bigl((X_\ell)_{\ell=0}^{L},(a_\ell)_{\ell=0}^{L-1}\bigr),
\end{equation}
where $X_\ell$ contains measured behavioral variables and $a_\ell$ constructs the next level. Proposed graphs $\widehat G_\ell$, selected relations $\mathcal R_\ell$, and hypothesis tests are associated with the levels where computed; a terminal graph is optional. All episode, coordinate and response spaces have specified sigma algebras, and all maps are measurable. Every coordinate is an explicit function of an episode. Graph edges themselves are estimates, so they become new variables only through these functions.

A hypothesis specifies a measured relation, outcome contrast, and scope. Descriptive hypotheses concern occurrence; predictive hypotheses concern later outcomes; interventional hypotheses concern feasible manipulations. Results are supported, refuted, or unresolved relative to the tests and assumptions. The search seeks abstractions that reduce unnecessary detail while preserving the contrasts required by $q$. Generalization is evaluated in held-out scenes, backbones, or their combinations; novelty is assessed against $K$.

\section{Recursive Trajectory Abstraction}
\label{sec:method}
\subsection{Event Measurements}
A measurement map $\phi_0$ yields $X_0=\phi_0(\tau)$. Each coordinate has an event family, agent role, temporal phase, support interval, and availability time. Families include communication, declared planning, tool/action execution, and observed context. Phases follow observable protocol boundaries. For predictive hypotheses, fix $t_{\rm pred}<t_{\rm target}$ and use coordinates available by $t_{\rm pred}$; feedback revealing an already realized target is excluded. Later coordinates support retrospective analysis. Final outcomes are not measurement inputs.

Executable predicates measure tool arguments and environmental events. Semantic predicates specify referents, coding rules, and independently assessed error. Each coordinate retains its categorical state and the support, referent, and eligibility metadata needed by later maps. Graph learning uses the categorical states; $x_B$ in a construction map includes their associated metadata. Every macro propagates the metadata required for further construction. Missing or ineligible events are recorded explicitly. For example, $D^k$ denotes detection of a discrepancy in phase $k$, $C^{k+1}$ a commitment referring to that discrepancy, and $R^{k+2}$ an action implementing its correction. The referent links distinguish one repair sequence from three unrelated events.

\subsection{Relation Proposal and Stability}
At level $\ell$, write $W=s(X_\ell)$ for categorical states and set $V_\ell=(A,W,Y)$. $A$ is one joint assignment node; its factors may be dependent. Graphs have no incoming edges to $A$, no outgoing edges from $Y$, and no backward temporal edges. Equal-phase relations require an explicit order or retain their orientation ambiguity. Acyclicity is enforced by the search.

One concrete choice is tier-constrained greedy equivalence search (GES) \citep{chickering2002optimal}. For categorical measurements, use a prespecified decomposable score such as
\begin{equation}
s(G)=\sum_{e\in\mathcal E_{\rm train}}
\bigl[-2\ell_e(G)+d_e(G)\log N_e\bigr],
\end{equation}
where $\ell_e$ is the maximized multinomial log likelihood, $d_e$ the parameter count, and $N_e$ the episode count. GES approximates score minimization; resampling or alternative scores assess sensitivity. This is a relation-proposal step. Causal interpretation requires the corresponding Markov, faithfulness, and confounding assumptions.

On independent screening episodes, compare each target conditional distribution $P_e(W_v\mid W_S=w_S)$ across environments. Retain a relation when simultaneous confidence bounds on the largest between-environment TV distance are below a prespecified margin $\delta_S$, over conditioning cells with sufficient support. Fix targets, $S$, margins, and the tested cell family using build data. Use cell-specific sample counts in simultaneous bounds; required cells with insufficient support remain unresolved. Adaptive screening beyond that fixed family is exploratory. Invariant causal prediction supplies identification results under its own assumptions \citep{peters2016invariant}; a generic stability pass instead nominates relations. Context-aware discovery is another option \citep{mooij2020joint}. Retain uncertainty in edge directions. Candidate blocks come from relations connected to $Y$ or a question-specific target; temporal precedence alone does not establish an effect \citep{runge2019detecting}.

\subsection{Motif Construction}
Let $\mathcal P_\ell$ contain proposed connected blocks of at least two behavioral coordinates, excluding $A,Y$. For each block $B$, select a function $f_B$ from a finite template family specified for the question. Templates include ordered conjunctions, counts, and categorical progress states. A binary repair motif is
\begin{equation}
M_{\rm repair}=D^kC^{k+1}R^{k+2},
\label{eq:repair}
\end{equation}
where the product uses binary state components on eligible episodes with the specified order and shared referent. Its value indicates completion of the defined sequence. An alternative categorical template distinguishes detection, commitment, and completed repair. Ineligible or missing cases retain their own category.

The map's support covers all required events; it becomes available at the last member's availability time. Assigned protocols and final outcomes are excluded from $f_B$. A name such as ``evidence-linked repair'' is attached after fixing the function. An LLM can propose predicates or names, but the resulting variable must remain computable and validated.

For a concrete selection rule, fit outcome predictors on build data before and after each proposed replacement. On screening data, retain the map when its increase in clipped log loss is at most a prespecified margin $\delta_Y$ in each environment and its target relations pass the stability criterion. This is a risk comparison between the fitted predictors. Select a disjoint collection $\mathcal B_\ell\subseteq\mathcal P_\ell$ in increasing description length, breaking ties by fixed coordinate order. Description length can rank maps that meet the same criteria \citep{grunwald2007mdl}; it is a compression preference rather than a causal test. Unmatched variables are carried forward. Writing $U_\ell=\{1,\ldots,p_\ell\}\setminus\bigcup_{B\in\mathcal B_\ell}B$ for their indices,
\begin{equation}
\begin{aligned}
a_\ell(A,x,Y)&=(A,x',Y),\\
x'&=\bigl((f_B(x_B))_{B\in\mathcal B_\ell},x_{U_\ell}\bigr).
\end{aligned}
\end{equation}
With $T_0(A,\tau,Y)=(A,\phi_0(\tau),Y)$, the recurrence is
\begin{equation}
T_{\ell+1}=a_\ell\circ T_\ell.
\end{equation}
Here $T_\ell(D)$ means rowwise application to the archive. Thus each next-level row contains measured motifs. In a constructed two-level example, $X_0=(D,C,R,H)$ describes discrepancy detection, referent-linked commitment, corrective action, and a subsequent handoff. The first map forms \mbox{$M=DCR$} and retains $H$, giving \mbox{$X_1=(M,H)$}. A second map forms \mbox{$J=MH$}, measured after the handoff and meaning ``completed repair followed by handoff.'' It is retained only if its specified predictive and intervention contrasts pass the tests. These products define motif occurrence for subsequent relational hypotheses.

\subsection{Intervention Validation}
A chosen implementation is a complete measurable policy $\kappa_j:h\mapsto\kappa(j,h)$ on available histories, including a specified action when no eligible event occurs: for example, insert a specified message, require verification before action, or change a role protocol. Its identity is assigned before the unit starts; later realized actions can depend on history. Compliance and collateral changes enter the resulting trajectory. The effect
\begin{equation}
\theta_e(j,j_0)=\E_e[Y^{\kappa_j}]-\E_e[Y^{\kappa_{j_0}}]
\label{eq:effect}
\end{equation}
concerns assignment to those protocols. Identifying a downstream behavior effect requires additional assumptions \citep{hernan2020whatif,robins2000marginal}.

\begin{figure}[tb]
\centering
\begin{tikzpicture}[font=\small,>=Latex,
 box/.style={draw=black!65,fill=black!4,rounded corners=2pt,align=center,text width=1.55cm,minimum height=.9cm,inner sep=4pt},
 candidate/.style={box,draw=academicblue,fill=academicblue!6,text width=1.55cm}]
\node[box] (p1) at (0,0) {Realization $i_1$\\$P_e^{i_1}$};
\node[box] (p2) at (0,-1.65) {Realization $i_2$\\$P_e^{i_2}$};
\node[box] (r1) at (2.7,0) {$R_1$\\states\\and $Y$};
\node[box] (r2) at (2.7,-1.65) {$R_2$\\states\\and $Y$};
\node[candidate] (q) at (5.45,-.825) {Candidate\\response\\$Q_e^j$};
\draw[->,thick] (p1) -- node[above,font=\footnotesize] {$\Phi_\ell^+$} (r1);
\draw[->,thick] (p2) -- node[above,font=\footnotesize] {$\Phi_\ell^+$} (r2);
\draw[<->,dashed,academicblue,thick] (r1.east) -- (q.north west);
\draw[<->,dashed,academicblue,thick] (r2.east) -- (q.south west);
\node[align=center,font=\footnotesize] at (2.5,-2.8) {$\omega(i_1)=\omega(i_2)=j$\\Maximum response error $\geq\tfrac12\TV(R_1,R_2)$};
\end{tikzpicture}
\caption{Different realizations of one abstract manipulation must support a common response. Disagreement rejects the common-response claim at the stated tolerance and motivates refinement or scope revision.}
\label{fig:fidelity}
\end{figure}
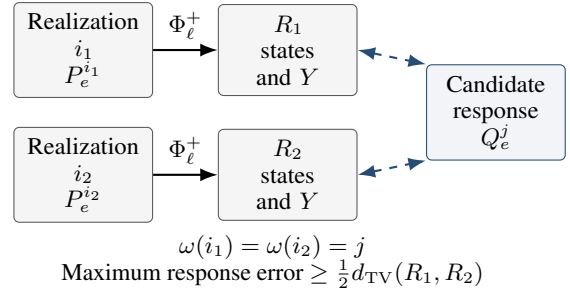

Let $P_e^i$ denote the joint assignment--trajectory--outcome law under implementation $i$, with $\omega(i)=j$, holding non-target factors fixed or using preassignment strata. The validation map is $\Phi_\ell^+=c_\ell\circ T_\ell$, where $c_\ell$ selects the scientific coordinates and outcome from a learning-table row. It excludes arm identifiers and construction-only timestamps or referent IDs. These remain metadata in the learning table $T_\ell$. An abstract response model $Q_{\ell,e}^j$ is fitted independently of confirmation data on the same selected measurable response space. Following causal abstraction \citep{rubenstein2017causal,beckers2020approximate}, compare
\begin{equation}
\eta_\ell=\max_{e,i}
\TV\bigl((\Phi_\ell^+)\push P_e^i,Q_{\ell,e}^{\omega(i)}\bigr)
\label{eq:fidelity}
\end{equation}
over a specified finite family of environments and implementations. Push-forward denotes applying the map to draws from the distribution. Implementation-specific hypotheses compare their documented protocol; implementation-independent hypotheses compare multiple realizations of the same $j$ (Figure~\ref{fig:fidelity}).

For a fixed categorical projection, such as $(M_B,Y)$, let $r\geq1$ be its number of states and $n\geq1$ iid confirmation units from that law, conditional on construction. The confidence allocation satisfies $0<\beta<1$. Coordinate-wise Hoeffding bounds give
\begin{equation}
b(n,r,\beta)=\min\left\{1,\frac r2
\sqrt{\frac{\log(2r/\beta)}{2n}}\right\}.
\end{equation}
With probability at least $1-\beta$, its empirical distribution $\widehat P$ satisfies \mbox{$\TV(P,\widehat P)\leq b$}. Hence $\TV(\widehat P,Q)+b\leq\epsilon$ certifies the projected tolerance for independently fitted $Q$. All tests, including rejected proposals and revisions, allocate confidence budgets $\beta_t$ with $\sum_t\beta_t\leq\beta$. The projection, map, response model, predictors, target laws, tolerances, batch size and allocation are fixed before confirmation; eligibility determined after assignment is retained as a category. Appendix~\ref{app:confidence} derives the bounds.

Confirm the jointly selected replacement, since separate block checks do not establish joint preservation. Predictors are fitted before this test; the upper confidence bound on the paired clipped-loss increase must not exceed $\delta_Y$ in every specified environment/arm or fixed target mixture. Failed or unresolved criteria trigger additional sampling, refinement, or restriction. Revised functions use fresh fixed-size confirmation batches and new confidence allocations. A layer is accepted only after all required criteria pass. Projected checks do not establish full-trajectory consistency.

\subsection{Recursion and Scientific Outputs}
Algorithm~\ref{alg:abstraction} specifies the construction. Each level uses independent build, screening, and confirmation data. An accepted map becomes the next-level table; lower levels remain available for comparison. The search stops when no reduction meets its criteria or resources are exhausted. Outputs retain accepted maps and hypotheses, including rejected and unresolved tests.

\begin{algorithm*}[t]
\caption{Recursive trajectory abstraction}
\label{alg:abstraction}
\begin{algorithmic}[1]
\raggedright
\Require Study settings; $\phi_0$; templates and tolerances
\State $T\gets T_0$; initialize $\mathcal H$ at $X_0$; $\ell\gets0$
\While{resources permit}
\State Acquire independent build and screening data
\State Fit $\widehat G_\ell$; screen relations; store test results
\State Select disjoint blocks and explicit $f_B$
\State \textbf{if} no qualifying reduction \textbf{then break}
\State Form $a_\ell$ with carry-forward; $T'\gets a_\ell\circ T$
\State Acquire independent confirmation episodes
\State Test and record joint predictive loss, responses, and transfer
\If{required criteria fail or remain unresolved}
\State Revise $a_\ell,T'$ or intervention scope
\State \textbf{if} resources insufficient \textbf{then break}
\State Acquire fresh fixed-size batch; allocate $\beta_t$; retest
\State \textbf{if} criteria still unmet \textbf{then break}
\EndIf
\State Append accepted maps and hypotheses to $\mathcal H$
\State $T\gets T'$; $\ell\gets\ell+1$
\EndWhile
\State Assess literature novelty and modeled belief change
\State \Return $\mathcal H$ and tested hypotheses
\end{algorithmic}
\end{algorithm*}

Render each hypothesis as a measured relation with its contrast and scope, then compare it with $K$ to assess novelty. Condition the prior and likelihood on completed exploration $\mathcal F$, including $K$. For \mbox{$\pi_h=P(h\mid\mathcal F)$}, independent confirmation defines a posterior $\pi_h'$ under the fixed hypothesis models. The surprise score $\operatorname{KL}(\operatorname{Ber}(\pi_h')\Vert\operatorname{Ber}(\pi_h))$ ranks model-relative belief change. Literature novelty is assessed separately; LLM belief elicitation is a heuristic unless the update is justified.

\section{Analytical Results}
\label{sec:analysis}
\subsection{Randomized Design and Finite Reduction}
\textbf{Proposition 1.} If $T_\ell$ is fixed independently of confirmation data, valid randomized assignment identifies its protocol effects. For a fixed bounded real statistic $V=h(X_\ell(\tau))$, such as the indicator of motif completion,
\begin{equation}
\begin{aligned}
&\E[V^a-V^{a'}\mid e]\\
&\quad=\E[V\mid A=a,e]-\E[V\mid A=a',e].
\end{aligned}
\end{equation}
This follows from assignment exchangeability, consistency, and positivity. It identifies the assigned protocol's effect on the measured abstraction regardless of graph correctness.

\textbf{Proposition 2.} Let $p_\ell=|X_\ell|$ with $p_0\geq1$. Disjoint block replacement with carry-forward gives
\begin{equation}
p_{\ell+1}=p_\ell-\sum_{B\in\mathcal B_\ell}(|B|-1).
\end{equation}
Every nonempty accepted reduction decreases $p_\ell$ by at least one. A fixed-inventory reduction chain therefore has at most $p_0-1$ transitions. Refinement, splitting, or feature additions outside that chain consume the research budget and are not covered by this bound.

\subsection{Realization Disagreement}
\textbf{Proposition 3.} If $i_1,i_2$ realize the same $j$ and $R_k$ is their abstract response distribution, then
\begin{equation}
\begin{aligned}
&\max_{k\in\{1,2\}}\TV(R_k,Q^j)\\
&\qquad\geq\tfrac12\TV(R_1,R_2).
\end{aligned}
\label{eq:disagree}
\end{equation}
The triangle inequality proves the bound. It rejects a common-response claim at insufficient tolerance; possible repairs include refining the block, implementations, response model, or scope.

In a constructed example, $X$ indicates a grounded acknowledgment, $W$ a generic acknowledgment, $Z=X+W$ their count, and $Y=X$ the stipulated outcome. The two protocols
\begin{equation}
\begin{aligned}
i_1&=\operatorname{do}(X=1,W=0),\\
i_2&=\operatorname{do}(X=0,W=1)
\end{aligned}
\end{equation}
both realize $j=$ deliver one acknowledgment, with $\omega(i_1)=\omega(i_2)=j$, but produce point masses at $(Z,Y)=(1,1)$ and $(1,0)$. Their TV distance is one. A common response at tolerance below $1/2$ is impossible. Distinguishing types requires refining the intervention labels; alternatively the hypothesis can be restricted to one protocol. The example specifies the outcome law; it is not an empirical claim about acknowledgments.

\subsection{Composition of Consistency Error}
\textbf{Proposition 4.} Fix an environment and preassignment stratum. Let $j_0=i$ and $j_{\ell+1}=\omega_\ell(j_\ell)$. On specified measurable response spaces, let $R_0^i=(\Phi_0^+)\push P_e^i$ be the actual initial law and $R_\ell^{j_\ell}=Q_{\ell,e}^{j_\ell}$ for $\ell\geq1$ be specified comparison laws. Suppose measurable outcome-preserving maps satisfy $c_{\ell+1}\circ a_\ell=\bar a_\ell\circ c_\ell$ and
\begin{equation}
\TV((\bar a_\ell)\push R_\ell^{j_\ell},R_{\ell+1}^{j_{\ell+1}})\leq\epsilon_\ell.
\end{equation}
Then, with $\Psi_L=\bar a_{L-1}\circ\cdots\circ\bar a_0$,
\begin{equation}
\begin{aligned}
&\TV\bigl((\Psi_L)\push R_0^i,R_L^{j_L}\bigr)\\
&\qquad\leq\min\Bigl\{1,\sum_{\ell=0}^{L-1}\epsilon_\ell\Bigr\}.
\end{aligned}
\label{eq:compose}
\end{equation}
TV contraction and induction with the triangle inequality prove the bound. The factorization requires projections retaining sufficient linkage and eligibility summaries. The first tolerance compares actual data to the first model; later tolerances compare specified model laws. The bound concerns the specified adjacent model comparisons. Appendix~\ref{app:proofs} gives the proof.

\section{Discussion}
The formulation specifies how measured motifs become objects of further scientific study. Its practical value should be evaluated against semantic-category induction, automata abstraction, and prediction-driven compression. Matched-budget comparisons can assess independently rejected hypotheses, appropriate unresolved conclusions, calibration, and transfer across scenes and backbones. High-dimensional conditioning and projected-response tests can require many episodes, motivating sparse templates, better experimental design \citep{myung2013adaptive}, and partial identification. Flat semantic coding is sufficient when a question concerns event prevalence or when no validated reduction is available. A hierarchy is useful when a measured motif supports a new relational hypothesis with fewer variables. Descriptive studies need observations, predictive studies need independent future-outcome tests, and causal hypotheses need identified interventions. Compression can remove precisely the distinction that matters to a hypothesis, so retaining a lower-level representation is a scientific outcome. The number of levels is therefore not a quality measure.

\section{Conclusion}
We formulate scientific discovery of language-agent behavior through a concrete recursive abstraction procedure. Explicit episode functions construct motifs from proposed relations; independent tests determine their retention, refinement, and scope. Randomized protocols anchor scientific contrasts, while realization disagreement and consistency bounds constrain the hierarchy. The formulation provides inputs, outputs, construction rules, validation tests, and stopping conditions for subsequent empirical research.

\section*{Limitations}
The method is proposed and analyzed, with a constructed example; it has not been implemented or empirically evaluated here. The narrative review is bounded by its publication cutoff and selection criteria. Template selection, semantic coding, phase alignment, and intervention feasibility remain study-specific.

GES and invariance procedures require model assumptions that may fail for language-agent trajectories or after coarsening. Hidden state, persistent memory, selection, and incomplete logs can affect identification. Stability and disjoint source coordinates do not establish causality or faithfulness. Protocol effects do not automatically identify the effects of downstream behavioral variables.

The finite-sample bound assumes fixed categorical projections and independent confirmation units; it can be conservative. Projected tests support only their measured contrasts. The finite-depth result concerns accepted fixed-inventory reductions, and the composition result requires matched layer-wise distributions. Neither establishes scientific completeness or inevitable improvement with depth. Generalization beyond tested conditions and literature-relative novelty require additional evidence.

\section*{Ethical Considerations}
Trajectory studies can expose private messages and tool outputs. Applications should minimize retained data, document consent and access where relevant, and prevent experimental protocols from causing unauthorized actions. Simulator findings require separate evidence for deployment and human behavior. The example is constructed and contains no human participant data. Human authors remain responsible for scientific claims and automated assistance.

\bibliography{refs}
\appendix

\section{Review Scope and Comparisons}
\label{app:review}
The review covers trajectory measurement, inductive behavioral analysis, causal abstraction, and automated scientific research. Searches of primary publication and preprint sources were complemented by backward and forward reference searches and verification of specific related works. Selection favored studies specifying measurements, construction procedures, experimental contrasts, or formal assumptions relevant to language-agent trajectories. The publication cutoff is 6 October 2026; topic-based searches on 7 October checked coverage against that cutoff. We distinguish peer-reviewed publications from author versions and preprints. The synthesis is a narrative review rather than an exhaustive systematic review.

\begin{table*}[t]
\centering
\begin{tabular}{@{}p{.16\textwidth}>{\raggedright\arraybackslash}p{.38\textwidth}>{\raggedright\arraybackslash}p{.40\textwidth}@{}}
\toprule
Work & Relevant construction & Relation to this formulation \\
\midrule
AutoTraceGT & Grounded coding, category/relation revision, saturation, theory generation. & Already specifies a study process; our target adds recursive measured motifs and protocol-response tests. \\
Actonomy & Hierarchical action vocabulary and trajectory coding. & Supplies event categories; hypothesis-specific experimental distinctions determine reductions. \\
ATLAS & Trace abstraction, concretization, automata learning, symbolic transfer. & Recovers transitions; our target is motif construction for tested behavioral hypotheses. \\
ReBeCA & Semantic patterns, graph learning, stability checks, prompt interventions. & Closely related operations; our formulation makes recursive maps and realization dependence explicit. \\
CAMO & Factor construction and paired simulator interventions. & Provides experimental factor analysis; our maps define next-level episode measurements. \\
COAT / IRIS & Variable proposal and relation discovery from unstructured evidence. & Motivates construction; our recursion preserves availability and tests feasible realizations. \\
\bottomrule
\end{tabular}
\caption{Selected trajectory-analysis methods and their relation to the proposed scientific target.}
\end{table*}

\section{Measurement Specification}
\label{app:measurement}
A coordinate is specified by its event predicate, role, phase, referent, support interval, availability time, codomain, and missingness rule. For the repair example, detection requires an observable discrepancy statement; commitment must identify that discrepancy and an intended correction; repair requires execution feedback matching the correction. A label inferred from an utterance is a measurement of expressed content. It does not identify latent intention.

An ordered conjunction requires the declared temporal order and shared referent. Let $\chi_B(x_B)$ indicate eligibility. Eligibility, temporal order, and shared referents are determined from the coordinate metadata retained at the current level. Write \mbox{$w_v=s_v(x_v)$} for each state component. The macro state is
\begin{equation}
 m_B(x_B)=\begin{cases}
 \prod_{v\in B}w_v,&\chi_B(x_B)=1,\\
 \bot,&\chi_B(x_B)=0.
 \end{cases}
\end{equation}
This equation defines the macro state. The complete record is $f_B(x_B)=(m_B(x_B),k_B(x_B))$, where $k_B$ retains the support, referent and eligibility summaries needed by subsequent maps. The symbol $\bot$ records ineligibility; missing observation can be another category. Successor maps access only the current coordinate records. Products and graph scores use state components; the projection used for a finite-state test excludes identifiers and timestamps. Prospective coordinates access only their available histories. Training may use outcome supervision to select a function, but the frozen function takes neither final outcome nor assigned factor as an input. The next learning table preserves those fields separately. Validation compares behavior and outcome coordinates, with non-target factors fixed or stratified; realization identifiers remain outside this response distribution.

Categorical alternatives can distinguish no detection, detection without commitment, commitment without completed correction, completed repair, and ineligibility. Choice of a binary or categorical map is a testable modeling decision: it changes both compression and the distinctions that later hypotheses can express. If role or phase boundaries differ across scenes, alignment is defined before comparison or the hypothesis scope is restricted.

On observed eligible phase windows, the constructed two-level example gives the state components
\begin{equation}
\begin{aligned}
 X_0&=(D,C,R,H),\\
 X_1&=(DCR,H)=(M,H),\\
 X_2&=MH=J.
\end{aligned}
\end{equation}
For instance, $(D,C,R,H)=(1,1,0,1)$ yields $(M,H)=(0,1)$ and $J=0$; $(1,1,1,1)$ yields $(1,1)$ and $J=1$. These values demonstrate the maps. 

\section{Proofs and Identification}
\label{app:proofs}
\subsection{Randomized Assignment}
Condition on completed construction, including the fixed measurement function and the unit-level design assumptions in Section~\ref{sec:problem}. Suppose assignment $A$ is randomized within environment $e$, with positive probability for each compared arm. For a fixed bounded statistic $V=h(X_\ell(\tau))$, consistency gives $V=V^a$ when $A=a$, and assignment exchangeability gives $V^a\perp A\mid e$. Therefore
\begin{equation}
\begin{aligned}
 \E[V\mid A=a,e]
 &=\E[V^a\mid A=a,e]\\
 &=\E[V^a\mid e].
\end{aligned}
\end{equation}
Subtracting the corresponding identity for $a'$ proves Proposition~1. Randomization identifies assignment effects, including noncompliance. A prompt requiring an action is an assigned protocol; replacing it with a perfect intervention on the resulting action requires additional identification assumptions.

Randomization of design factors does not orient every relation between behavioral variables. Let $U$ be a fair binary variable independent of an assigned factor. Models $X=U,Y=U$ and $X=U,Y=X$ have identical observational distributions. Setting $X$ to zero leaves the first outcome random and fixes the second at zero. Thus behavioral causal effects remain unidentified from these observations, despite randomized assignment of an unrelated factor.

\subsection{Reduction Depth}
For disjoint blocks, replacing $|B|$ coordinates by one removes $|B|-1$ coordinates. All unmatched coordinates remain, proving the exact recurrence in Proposition~2. Each block has size at least two, so each accepted transition decreases the positive integer $p_\ell$. Hence there are at most $p_0-1$ transitions in a fixed-inventory chain. Retaining lower-level measurements permits later scientific comparison without counting them as new coordinates within that chain.

If blocks partition all coordinates and every block has size at least two, then $p_{\ell+1}\leq\lfloor p_\ell/2\rfloor$, yielding at most $\lfloor\log_2 p_0\rfloor$ transitions. This stronger bound does not apply to carry-forward reductions. Introducing features, splitting variables, or rolling back restarts the relevant chain; the research budget bounds those operations separately.

\subsection{Realization Dependence}
With \mbox{$d_k=\TV(R_k,Q)$} for a common response $Q$, the triangle inequality gives
\begin{equation}
\begin{aligned}
 \TV(R_1,R_2)
 &\leq d_1+d_2\\
 &\leq2\max_k d_k.
\end{aligned}
\end{equation}
This proves Proposition~3. The constructed acknowledgment example has disjoint outcome support under the two protocols, so its TV distance equals one. Its repair options are to retain the distinction between grounded and generic messages or to make an implementation-specific claim.

\subsection{Composition}
Use the fixed environment, response spaces, comparison laws and factorization in Proposition~4. Let $\Psi_0$ be the identity and $\Psi_\ell=\bar a_{\ell-1}\circ\cdots\circ\bar a_0$. The assumed layer comparison is
\begin{equation}
\TV((\bar a_\ell)\push R_\ell^{j_\ell},R_{\ell+1}^{j_{\ell+1}})\leq\epsilon_\ell.
\end{equation}
For a measurable map, TV contracts because each output event has a measurable inverse image. Insert $(\bar a_\ell)\push R_\ell^{j_\ell}$ as an intermediate law:
\begin{equation}
\begin{aligned}
&\TV((\Psi_{\ell+1})\push R_0^i,R_{\ell+1}^{j_{\ell+1}})\\
&\quad\leq\TV((\Psi_\ell)\push R_0^i,R_\ell^{j_\ell})+\epsilon_\ell.
\end{aligned}
\end{equation}
The initial discrepancy is zero. Induction proves the sum bound, capped at one.  If two records have identical selected states but differ in linkage, the next motif may be one versus $\bot$. A projection discarding that distinction has no induced map; its comparisons remain level-specific. A common abstract label also does not define a unique actual law when realizations disagree. The higher-level $R$ laws here are explicitly chosen comparison models.

An invariance test's retention event does not automatically establish causal correctness. If a discovery procedure has a justified conditional error guarantee $\alpha_\ell$ at each adaptively chosen level, a union bound controls the probability of any corresponding error by $\sum_\ell\alpha_\ell$. Such a guarantee must be proved for that procedure and its post-abstraction assumptions. Independent data alone do not supply it.

\section{Finite-Sample Response Tests}
\label{app:confidence}
Condition on all pretest information. Let $P$ be the fixed law of the declared $r$-state projection, with $r\geq1$, $n\geq1$, and $0<\beta<1$. For state $v$, empirical frequency $\widehat p_v$ from $n$ iid experimental units drawn from $P$ obeys Hoeffding's inequality \citep{hoeffding1963probability}:
\begin{equation}
 P(|\widehat p_v-p_v|>t)\leq2e^{-2nt^2}.
\end{equation}
Taking a union bound over states with $t=\sqrt{\log(2r/\beta)/(2n)}$ yields
\begin{equation}
 \TV(P,\widehat P)=\tfrac12\sum_v|p_v-\widehat p_v|
 \leq b(n,r,\beta)
\end{equation}
with probability at least $1-\beta$. Clipping at one is valid because TV never exceeds one.

For predictive confirmation, use two fixed predictors and clipped losses in $[0,C]$, with $C>0$ fixed before testing. On the same iid units let $\Delta_u=\ell_{\rm new,u}-\ell_{\rm old,u}$ and $\widehat\Delta=n^{-1}\sum_u\Delta_u$. Since $\Delta_u\in[-C,C]$, Hoeffding bounds the expected increase by
\begin{equation}
\E\Delta_u\leq\widehat\Delta+C\sqrt{\frac{2\log(1/\beta_t)}{n}}
\end{equation}
with probability at least $1-\beta_t$. Accept only if this upper bound is at most $\delta_Y$, for the joint replacement in every required target law. The criterion concerns the expected clipped-risk contrast of the fixed predictors.

For a response model $Q$ fitted independently of the confirmation batch, the triangle inequality gives
\begin{equation}
 \TV(P,Q)\leq\TV(\widehat P,Q)+b(n,r,\beta).
\end{equation}
For two independently sampled realizations with confidence allocations $\beta_1+\beta_2\leq\beta$, put $b_k=b(n_k,r,\beta_k)$. Let $\widehat d=\TV(\widehat P_1,\widehat P_2)$. With probability at least $1-\beta$,
\begin{equation}
\begin{aligned}
 \TV(P_1,P_2)&\geq\max\{0,\widehat d-b_1-b_2\},\\
 \TV(P_1,P_2)&\leq\min\{1,\widehat d+b_1+b_2\}.
\end{aligned}
\end{equation}
If half the lower bound exceeds a proposed common-response tolerance, Proposition~3 rejects that response scope. If the upper bounds to independently fitted $Q$ fall below the tolerance for all required arms, the projected criterion is met. Otherwise the conclusion is unresolved.

Before attempt $t$, the completed history $\mathcal F_{t-1}$ fixes maps, projections, models, predictors, target laws, tolerances, sample sizes and allocations. Each fresh batch is conditionally iid from its declared law. Split $\beta_t$ among every tested arm, environment, cell and quantity, or supply a simultaneous guarantee. If failure event $E_t^c$ has conditional probability at most $\beta_t$ and $\sum_t\beta_t\leq\beta$ almost surely, then
\begin{equation}
\Pr(\exists t:E_t^c)\leq\E\sum_t\beta_t\leq\beta.
\end{equation}
Every confirmation attempt uses a fresh fixed-size batch. Completed batches enter the construction history for subsequent independently confirmed proposals. For clusters, this simple proof uses one prespecified finite-valued observation per iid cluster. Episode-weighted cluster analyses specify their corresponding sampling bound separately. A post-assignment missing or ineligible state remains in the distribution; removing it can change the protocol comparison. Tests on $(M_B,Y)$ establish that projected comparison, not equality of unrestricted text-trajectory laws.

\section{Novelty and Future Evaluation}
A hypothesis is represented by its measured variables, claimed relation, contrast, and scope. Comparison with existing work distinguishes replication, scope extension, and previously untested relations. The literature search and its date delimit an underexploration claim. Supported known hypotheses remain legitimate scientific findings.

For a selected hypothesis, condition on completed exploration $\mathcal F$. Specify exhaustive alternative models $h,\neg h$, a prior $0<\pi_h=P(h\mid\mathcal F)<1$, and predictive likelihoods $L_h=P(D_c\mid h,\mathcal F)$ and $L_0=P(D_c\mid\neg h,\mathcal F)$ for the independent batch under the fixed design. Composite models integrate nuisance parameters under their specified priors. If $\pi_h L_h+(1-\pi_h)L_0$ is positive and finite, then
\begin{equation}
 \pi_h'=\frac{\pi_h L_h}{\pi_h L_h+(1-\pi_h)L_0}.
\end{equation}
Use $0\log0=0$ for posterior endpoints. The KL score measures change within that model. It can prioritize investigation but does not replace experimental support or literature comparison.

Future evaluation should compare the proposed construction with fixed action taxonomies, grounded-theory induction, automata learning, and prediction-only compression at matched data and experimental budgets. Controlled tasks with stipulated mechanisms can assess abstraction recovery and appropriate rejection. Held-out scenes and backbones test transfer of measured relations and intervention responses. Required analyses include coding error, eligibility rules, role/phase alignment, template choice, graph-score assumptions, independent data partitions, and calibration of unresolved conclusions. 
\end{document}